\documentclass[runningheads]{llncs}
\usepackage[T1]{fontenc}
\usepackage{graphicx}
\usepackage{microtype}
\usepackage{booktabs}
\usepackage{url}
\begin{document}
\title{FedOUI: OUI-Guided Client Weighting \\ for Federated Aggregation}
%
%
%

\author{
Alberto Fernández-Hernández\inst{1} \and
Jose I. Mestre\inst{1} \and
Cristian Pérez-Corral\inst{1} \and
Manuel F. Dolz\inst{2} \and
Jose Duato\inst{3} \and
Enrique S. Quintana-Ortí\inst{1}
}

\authorrunning{A. Fernández-Hernández et al.}

\institute{
Universitat Politècnica de València, Valencia, Spain\\
\email{\{a.fernandez,jimesmir,cpercor\}@upv.es, quintana@disca.upv.es}
\and
Universitat Jaume I, Castelló de la Plana, Spain 
\email{dolzm@uji.es}
\and
Openchip \& Software Technologies S.L., Spain 
\email{jose.duato@openchip.com}
}

%
\maketitle              
\begin{abstract}
Federated learning usually aggregates client updates using dataset size or gradient-level criteria, while overlooking internal signals about how each client model is organizing its input space during training. We introduce \emph{FedOUI}, a simple aggregation rule based on the Overfitting--Underfitting Indicator (OUI), an activation-based and label-free metric. Each participating client sends its local update together with a OUI value computed on a fixed probe batch, and the server estimates the round-wise OUI distribution to assign lower weights to structurally atypical clients through a smooth reweighting rule. We evaluate FedOUI on CIFAR-10 under strong non-IID partitioning and noisy-client conditions, comparing it with FedAvg, FedProx, and a gradient-alignment baseline. The clearest gains appear under strong heterogeneity, where OUI-based weighting improves aggregation quality while remaining lightweight and interpretable. These results show that internal activation structure can provide useful information for federated aggregation beyond client size and gradient geometry.

\keywords{Overfitting-Underfitting Indicator, Federated learning, activation-based weighting, client aggregation}
\end{abstract}
\section{Introduction}

Federated learning (FL)~\cite{konecny_federated_2016} trains a shared model across distributed clients without centralizing raw data. Its main challenge is heterogeneity: clients differ in sample size, class distribution, noise level, and local optimization dynamics, so their updates carry different value for the global model. Standard aggregation rules only capture part of this variation. Standard aggregation rules~\cite{mcmahan_communication-efficient_2017} only capture part of this variation. Later methods address heterogeneity~\cite{li_federated_2020,mendieta_local_2022}, improve robustness through robust aggregation~\cite{pillutla_robust_2022}, or use gradient similarity and client coherence to adapt aggregation~\cite{li_revisiting_2023}, thus focusing on the updates themselves.

This paper adds a complementary signal. We build on the Overfitting--Underfitting Indicator (OUI)~\cite{fernandez-hernandez_oui_2025}, an activation-based metric that summarizes how balanced the activation patterns of a layer are over a probe batch. Recent results~\cite{fernandez-hernandez_when_2026} show that OUI acts as an early structural signal of training regime and can be monitored online at low cost. This makes it a natural candidate for FL: clients participating in the same round may produce updates of similar scale while operating in very different internal regimes.

We propose \emph{FedOUI}, a server-side aggregation rule in which each client sends its local update together with an OUI value computed on a fixed probe batch. At each round, the server estimates the distribution of the received OUI values and softly down-weights clients in the tails, interpreting OUI as a structural typicality signal. The resulting method is simple, compatible with standard synchronous FL, and easy to inspect.

We evaluate FedOUI on CIFAR-10 with a small CNN in two controlled settings: strong non-IID partitioning and noisy clients. We compare FedOUI against standard aggregation and heterogeneity-aware baselines, namely FedAvg~\cite{mcmahan_communication-efficient_2017}, FedProx~\cite{li_federated_2020}, and FedAlign~\cite{mendieta_local_2022}. The main empirical pattern is clear: OUI-based weighting is most useful under strong heterogeneity, where it improves aggregation quality over the standard baselines, while in noisier settings it remains competitive. These results position activation structure as a practical source of information for federated aggregation.

\section{Related Work}

Federated learning is commonly framed around the aggregation of client updates under data heterogeneity. FedAvg~\cite{mcmahan_communication-efficient_2017} established the standard formulation, where the server averages local updates with weights proportional to client data size. This remains the main reference point because it is simple, efficient, and effective in many practical settings.

A second line of work focuses on stabilizing training under heterogeneous clients. FedProx~\cite{li_federated_2020} introduces a proximal term in the local objective to reduce client drift and improve optimization when local data distributions differ substantially. This direction addresses instability at the client level and is often used as a stronger baseline than FedAvg in non-IID settings.

A third line studies how the server should weight or aggregate updates once they are received. Robust aggregation methods, such as RFA ~\cite{pillutla_robust_2022}, replace the standard mean with more outlier-resistant rules based on geometric medians. Related work on weighted aggregation revisits the role of client importance directly and shows that aggregation weights can influence both optimization and generalization beyond raw sample count alone~\cite{li_revisiting_2023}. These methods motivate a broader view of aggregation in which clients contribute unequally according to signals that reflect their usefulness for the global model.

Our method is closest to this last family, but it uses a different signal. Instead of weighting clients only from update geometry or robustness considerations, we use OUI as a compact activation-based measure of structural typicality. This places FedOUI within the literature on adaptive client weighting while introducing an internal, label-free signal derived from the client model itself.

\section{Method}

We consider synchronous federated learning with a central server and a subset of clients sampled at each communication round. At round $t$, each selected client $k$ starts from the current global model, performs local training on its private data, and returns three quantities to the server: its local model update $\Delta_k^t$, its local sample count $n_k$, and a OUI value $o_k^t$. The server then computes aggregation weights from these quantities and updates the global model through a weighted average.

The OUI value is computed on the client side from the penultimate pre-activation layer using a fixed local probe batch of size $B$. Let $a_j(x_b;\theta_t)$ be the pre-activation of unit $j$ for sample $x_b$ under the local model parameters. We define the binary activation mask
\begin{equation}
m_{b,j}(t)=\mathbf{1}\{a_j(x_b;\theta_t)>0\},
\end{equation}
the number of active samples for unit $j$,
\begin{equation}
s_j(t)=\sum_{b=1}^{B} m_{b,j}(t),
\end{equation}
and the corresponding minority count,
\begin{equation}
u_j(t)=\min\!\left(s_j(t),\, B-s_j(t)\right).
\end{equation}
For a layer with $d$ units, the client OUI is
\begin{equation}
o_k^t=
\frac{1}{d}
\sum_{j=1}^{d}
\frac{u_j(t)}{\lfloor B/2 \rfloor}.
\end{equation}
This produces a scalar in $[0,1]$ that summarizes how balanced the activation pattern is over the probe batch. In our method, the OUI value is interpreted through its relative position within the set of clients participating in the round.

\medskip

Given the OUI values $\{o_k^t\}$ received at round $t$, the server needs a simple way to tell which clients lie near the center of the round and which ones lie toward the extremes. Since OUI is a bounded scalar in $[0,1]$, we model its round-wise distribution with a Beta law,
\begin{equation}
o_k^t \sim \mathrm{Beta}(\alpha_t,\beta_t),
\end{equation}
which is a flexible distribution defined on the interval $[0,1]$ and can capture both concentrated and skewed shapes. This fitted distribution gives a round-specific notion of structural typicality.

We then assign each client a bilateral structural score that is high near the middle of the round-wise OUI distribution and decreases smoothly toward both tails. Let $F_t$ denote the cumulative distribution function of the fitted Beta model at round $t$. The score is defined as
\begin{equation}
s_k^t = 2 \min \left\{ F_t(o_k^t),\, 1-F_t(o_k^t) \right\}.
\end{equation}
With this definition, clients close to the central mass of the OUI distribution receive larger scores, while clients in either tail receive smaller scores.

The main method, \emph{FedOUI}, combines local sample size and structural score through
\begin{equation}
w_k^t \propto n_k \, (\varepsilon + s_k^t),
\end{equation}
where $\varepsilon$ is a small constant for numerical stability, set to $10^{-3}$ in all experiments. The weights are then normalized so that $\sum_k w_k^t = 1$. The global update is then
\begin{equation}
\Delta^t = \sum_k w_k^t \Delta_k^t.
\end{equation}
This yields a soft reweighting rule in which clients with atypical values of OUI contribute less to the aggregation, while all selected clients remain active.

\section{Empirical Evaluation}
This section presents the empirical evaluation of FedOUI. We first describe the common experimental setup and then report the results in the two retained scenarios, followed by a representative round-level analysis of the OUI-based weighting mechanism.

\subsection{Experimental Setup}
We evaluate the proposed method on CIFAR-10~\cite{krizhevsky_learning_2009} using a small CNN. The model takes RGB images as input and consists of two convolutional blocks, \texttt{Conv2d(3, 32, kernel=3, padding=1)} and \texttt{Conv2d(32, 64, kernel=3, padding=1)}, \linebreak each followed by ReLU and \texttt{MaxPool2d(2,2)}. The convolutional backbone is followed by a fully connected head with \texttt{Linear(64*8*8, 128)}, ReLU, and \texttt{Linear(128, num\_classes)}. The federated setting uses 20 clients, with 5 clients sampled at each communication round. Each selected client performs one local epoch with SGD, momentum $0.9$, learning rate $0.01$, and batch size $32$. We run all experiments for 60 rounds and report results over three random seeds. For OUI computation, each client uses a fixed probe batch of size $32$, and the metric is computed on the penultimate pre-activation, that is, on the 128-dimensional vector before the last hidden-layer ReLU.

We compare FedOUI against FedAvg, FedProx, and FedAlign. All runs use a train subset of 3000 samples and a test subset of 1000 samples. The exact experiment configurations are provided through \url{https://github.com/anonymous/fedoui}.

We report three primary metrics: final test accuracy, best test accuracy reached during training, and the area under the test-accuracy curve (AUC)~\cite{huang_using_2005}. As secondary diagnostics, we inspect the round-wise OUI histograms, representative client-weight snapshots, and the fitted Beta parameters used by the structural score. These quantities are used to illustrate the behavior of the method rather than as standalone selection criteria.

\medskip
We now present the results in two evaluation scenarios and then examine a representative round to illustrate the structural weighting induced by FedOUI.

\subsection{Strong Non-IID Dirichlet Setting}

The main experiment uses a strong non-IID partition generated through a Dirichlet split with concentration parameter $0.1$. 
In this case, FedOUI achieves the best final accuracy and the best peak accuracy among the compared methods. This is the central empirical result of the paper. FedAvg and FedProx remain competitive, but both stay below FedOUI, while FedAlign is clearly weaker in this regime.

\begin{table}[ht]
\centering
\caption{Results in the strong non-IID Dirichlet setting. Mean $\pm$ standard deviation over 3 seeds.}
\label{tab:results_dirichlet}
\renewcommand{\arraystretch}{1.15}
\begin{tabular*}{\linewidth}{@{\extracolsep{\fill}}lccc@{}}
\toprule
\textbf{Method} & \textbf{Final accuracy} & \textbf{Best accuracy} & \textbf{Accuracy AUC} \\
\midrule
FedAlign & $0.1640 \pm 0.0234$ & $0.1937 \pm 0.0096$ & $0.1282 \pm 0.0049$ \\
FedAvg   & $0.2163 \pm 0.0244$ & $0.2797 \pm 0.0378$ & $0.1644 \pm 0.0089$ \\
FedProx  & $0.2130 \pm 0.0180$ & $0.2773 \pm 0.0393$ & $0.1647 \pm 0.0090$ \\
FedOUI   & $\mathbf{0.2343 \pm 0.0631}$ & $\mathbf{0.2820 \pm 0.0266}$ & $\mathbf{0.1726 \pm 0.0136}$ \\
\bottomrule
\end{tabular*}
\end{table}

\subsection{Noisy-Client Setting}

The secondary experiment introduces noisy clients to evaluate the method outside its most favorable regime. Here, FedOUI achieves the highest best accuracy, showing that structural weighting remains effective when client quality is degraded. Final accuracy is slightly lower than for FedAvg and FedProx, while the AUC remains close to both baselines. Overall, this result shows that FedOUI stays competitive under noisy clients and preserves a clear advantage in peak performance, which is the most relevant signal in this setting.

\begin{table}[ht]
\centering
\caption{Results in the noisy-client setting. Mean $\pm$ standard deviation over 3 seeds.}
\label{tab:results_noisy}
\renewcommand{\arraystretch}{1.15}
\begin{tabular*}{\linewidth}{@{\extracolsep{\fill}}lccc@{}}
\toprule
\textbf{Method} & \textbf{Final accuracy} & \textbf{Best accuracy} & \textbf{Accuracy AUC} \\
\midrule
FedAlign & $0.2270 \pm 0.0669$ & $0.2927 \pm 0.0237$ & $0.1816 \pm 0.0120$ \\
FedAvg   & $\mathbf{0.2727 \pm 0.0335}$ & $0.3100 \pm 0.0128$ & $\mathbf{0.2097 \pm 0.0164}$ \\
FedProx  & $\mathbf{0.2727 \pm 0.0309}$ & $0.3100 \pm 0.0101$ & $0.2093 \pm 0.0162$ \\
FedOUI   & $0.2310 \pm 0.0527$ & $\mathbf{0.3183 \pm 0.0090}$ & $0.2061 \pm 0.0155$ \\
\bottomrule
\end{tabular*}
\end{table}

\subsection{Round-Level OUI Distribution and Structural Weighting}

The round-level analysis matches the intended mechanism. In the representative Dirichlet round, the OUI values are centered around $0.2767$ and are well described by a Beta distribution with parameters $\alpha=6.10$ and $\beta=15.94$. The highest weight is assigned to the client whose OUI is closest to the center of the round distribution, while lower and higher OUI values receive smaller but still nonzero weights. This is the expected behavior of a soft structural weighting rule: central clients are emphasized and extremes are smoothly attenuated.

\begin{figure}[h]
    \centering
    \includegraphics[width=\linewidth]{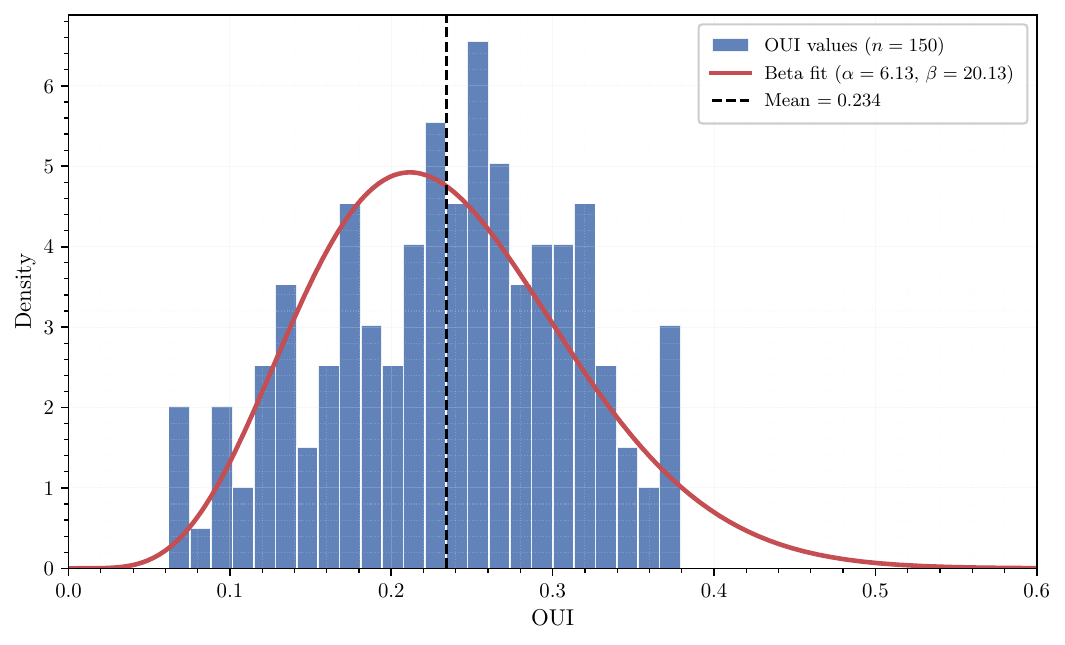}
    \caption{Representative round-wise OUI distribution in the strong non-IID setting, together with the fitted Beta density. The fit is stable and defines a meaningful central region for structural weighting.}
    \label{fig:oui_histogram}
\end{figure}

\section{Conclusions and future work}

FedOUI provides a simple and effective way to incorporate activation structure into federated aggregation. Across the experiments, it behaves as a soft structural weighting rule: clients near the center of the round-wise OUI distribution receive greater influence, while more atypical clients are smoothly attenuated. This mechanism is especially effective under strong non-IID heterogeneity, where FedOUI improves over FedAvg and FedProx, and it also remains robust in the noisy-client setting, where it achieves the strongest peak accuracy.

The empirical picture is clear. OUI captures a useful notion of client typicality, and that signal can guide aggregation with very little additional machinery. This positions activation-based observables as a practical ingredient for federated learning, extending their role from training analysis to aggregation control.

These results open a broad and promising research direction. A natural next step is to evaluate FedOUI on larger datasets, stronger architectures, and a wider range of federated settings, including additional forms of statistical heterogeneity, label corruption, and client imbalance. This would clarify how the structural signal evolves across scales and how its benefits transfer across tasks.

A second direction is methodological refinement. The current formulation already shows that round-wise structural typicality is useful, and future versions can enrich this idea through layer-wise OUI, temporal smoothing across rounds, or adaptive combinations with gradient geometry and trust signals. This could yield richer aggregation rules while preserving the simplicity of the current approach.

A third direction is systems integration. FedOUI offers a compact server-side mechanism, which makes it a good candidate for privacy-aware implementations, communication-efficient variants, and broader federated benchmarks. In parallel, the activation-centric view behind the method suggests a wider agenda: internal structural signals may become a valuable family of tools for monitoring, interpreting, and controlling distributed learning dynamics.

\section*{Acknowledgements}
This research was funded by the projects PID2023-146569NB-C21 and PID2023-146569NB-C22 supported by MICIU/AEI/10.13039/501100011033 and ERDF/UE. Alberto Fernández-Hernández was supported by the predoctoral grant PREP2023-001826 supported by MICIU/AEI/10.13039/501100011033 and ESF+. Jose I. Mestre was supported by the predoctoral grant ACIF/2021/281 of the \emph{Generalitat Valenciana}. Cristian Pérez-Corral received support from the \textit{Conselleria de Educación, Cultura, Universidades y Empleo} (reference CIACIF/2024/412) through the European Social Fund Plus 2021–2027 (FSE+) program of the \textit{Comunitat Valenciana}. Manuel F. Dolz was supported by grant {\small CNS2025-165098} funded by {\small MICIU/AEI/10.13039/501100011033} and by the Plan Gen--T grant {\small CIDEXG/2022/013} of the \emph{Generalitat Valenciana}.

%
%
%
\bibliographystyle{splncs04}
\bibliography{references}

@inproceedings{mcmahan_communication-efficient_2017,
    title = {Communication-{Efficient} {Learning} of {Deep} {Networks} from {Decentralized} {Data}},
    issn = {2640-3498},
    url = {https://proceedings.mlr.press/v54/mcmahan17a.html},
    language = {en},
    urldate = {2026-04-21},
    booktitle = {Proceedings of the 20th {International} {Conference} on {Artificial} {Intelligence} and {Statistics}},
    publisher = {PMLR},
    author = {McMahan, Brendan and Moore, Eider and Ramage, Daniel and Hampson, Seth and Arcas, Blaise Aguera y},
    month = apr,
    year = {2017},
    pages = {1273--1282},
}

@article{li_federated_2020,
    title = {Federated {Optimization} in {Heterogeneous} {Networks}},
    volume = {2},
    url = {https://proceedings.mlsys.org/paper/2020/hash/1f5fe83998a09396ebe6477d9475ba0c-Abstract.html},
    language = {en},
    urldate = {2026-04-21},
    journal = {Proceedings of Machine Learning and Systems},
    author = {Li, Tian and Sahu, Anit Kumar and Zaheer, Manzil and Sanjabi, Maziar and Talwalkar, Ameet and Smith, Virginia},
    month = mar,
    year = {2020},
    pages = {429--450},
}

@article{pillutla_robust_2022,
    title = {Robust {Aggregation} for {Federated} {Learning}},
    volume = {70},
    issn = {1053-587X, 1941-0476},
    url = {http://arxiv.org/abs/1912.13445},
    doi = {10.1109/TSP.2022.3153135},
    urldate = {2026-04-21},
    journal = {IEEE Transactions on Signal Processing},
    author = {Pillutla, Krishna and Kakade, Sham M. and Harchaoui, Zaid},
    year = {2022},
    note = {arXiv:1912.13445 [stat]},
    pages = {1142--1154},
}

@inproceedings{fernandez-hernandez_oui_2025,
    title = {{OUI} {Need} to {Talk} {About} {Weight} {Decay}: {A} {New} {Perspective} on {Overfitting} {Detection}},
    shorttitle = {{OUI} {Need} to {Talk} {About} {Weight} {Decay}},
    url = {https://ieeexplore.ieee.org/document/11159348},
    doi = {10.1109/AMLDS63918.2025.11159348},
    urldate = {2026-04-02},
    booktitle = {2025 {International} {Conference} on {Advanced} {Machine} {Learning} and {Data} {Science} ({AMLDS})},
    author = {Fernández-Hernández, Alberto and Mestre, Jose I. and Dolz, Manuel F. and Duato, Jose and Quintana-Ortí, Enrique S.},
    month = jul,
    year = {2025},
    pages = {96--105},
}

@misc{fernandez-hernandez_when_2026,
    title = {When {Learning} {Rates} {Go} {Wrong}: {Early} {Structural} {Signals} in {PPO} {Actor}-{Critic}},
    shorttitle = {When {Learning} {Rates} {Go} {Wrong}},
    url = {https://arxiv.org/abs/2603.09950v1},
    language = {en},
    urldate = {2026-04-13},
    journal = {arXiv.org},
    author = {Fernández-Hernández, Alberto and Pérez-Corral, Cristian and Mestre, Jose I. and Dolz, Manuel F. and Duato, Jose and Quintana-Ortí, Enrique S.},
    month = mar,
    year = {2026},
}

@article{konecny_federated_2016,
    title = {Federated {Optimization}: {Distributed} {Machine} {Learning} for {On}-{Device} {Intelligence}},
    shorttitle = {Federated {Optimization}},
    url = {https://www.semanticscholar.org/paper/Federated-Optimization%3A-Distributed-Machine-for-Konecn%C3%BD-McMahan/561269a24f2f2a06409109723a8ab93a01696efc},
    urldate = {2026-04-21},
    journal = {ArXiv},
    author = {Konecný, Jakub and McMahan, H. B. and Ramage, Daniel and Richtárik, Peter},
    month = oct,
    year = {2016},
}

@misc{li_revisiting_2023,
    title = {Revisiting {Weighted} {Aggregation} in {Federated} {Learning} with {Neural} {Networks}},
    url = {https://arxiv.org/abs/2302.10911v4},
    language = {en},
    urldate = {2026-04-21},
    journal = {arXiv.org},
    author = {Li, Zexi and Lin, Tao and Shang, Xinyi and Wu, Chao},
    month = feb,
    year = {2023},
}

@phdthesis{krizhevsky_learning_2009,
    type = {Technical {Report}},
    title = {Learning {Multiple} {Layers} of {Features} from {Tiny} {Images}},
    url = {https://www.cs.toronto.edu/~kriz/learning-features-2009-TR.pdf},
    school = {University of Toronto},
    author = {Krizhevsky, Alex},
    year = {2009},
}

@inproceedings{mendieta_local_2022,
    address = {New Orleans, LA, USA},
    title = {Local {Learning} {Matters}: {Rethinking} {Data} {Heterogeneity} in {Federated} {Learning}},
    copyright = {https://doi.org/10.15223/policy-029},
    isbn = {978-1-6654-6946-3},
    shorttitle = {Local {Learning} {Matters}},
    url = {https://ieeexplore.ieee.org/document/9880310/},
    doi = {10.1109/CVPR52688.2022.00821},
    language = {en},
    urldate = {2026-04-21},
    booktitle = {2022 {IEEE}/{CVF} {Conference} on {Computer} {Vision} and {Pattern} {Recognition} ({CVPR})},
    publisher = {IEEE},
    author = {Mendieta, Matias and Yang, Taojiannan and Wang, Pu and Lee, Minwoo and Ding, Zhengming and Chen, Chen},
    month = jun,
    year = {2022},
    pages = {8387--8396},
}

@article{huang_using_2005,
    title = {Using {AUC} and accuracy in evaluating learning algorithms},
    volume = {17},
    number = {3},
    journal = {IEEE Transactions on knowledge and Data Engineering},
    publisher = {IEEE},
    author = {Huang, Jin and Ling, Charles X},
    year = {2005},
    pages = {299--310},
}

\end{document}